\documentclass{article}
\usepackage{ijcai18}

\usepackage{times}
\usepackage{xcolor}
\usepackage{soul}
\usepackage[utf8]{inputenc}
\usepackage[small]{caption}

\usepackage{url}  
\usepackage{graphicx}  
\usepackage{latexsym}
\usepackage{amsmath,amssymb}
\usepackage{rotating}
\usepackage{subcaption}

\allowdisplaybreaks

\newif\ifappendix
\appendixtrue 

\title{Learning Conceptual Space Representations of Interrelated Concepts}

\author{Zied Bouraoui\\
CRIL CNRS \& Univ Artois, France\\
bouraoui@cril.univ-artois.fr\\
\And
Steven Schockaert\\
Cardiff University, UK\\
SchockaertS1@Cardiff.ac.uk
}
 
\begin{document}

\maketitle

\begin{abstract}
Several recently proposed methods aim to learn conceptual space representations from large text collections. These learned representations associate each \emph{object} from a given domain of interest with a point in a high-dimensional Euclidean space, but they do not model the \emph{concepts} from this domain, and can thus not directly be used for categorization and related cognitive tasks. A natural solution is to represent concepts as Gaussians, learned from the representations of their instances, but this can only be reliably done if sufficiently many instances are given, which is often not the case. In this paper, we introduce a Bayesian model which addresses this problem by constructing informative priors from background knowledge about how the concepts of interest are interrelated with each other. We show that this leads to substantially better predictions in a knowledge base completion task.
\end{abstract}

\section{Introduction}
Conceptual spaces are geometric representations of knowledge, in which the objects from some domain of interest are represented as points in a metric space, and concepts are modelled as (possibly vague) convex regions \cite{Gardenfors:conceptualSpaces}. The theory of conceptual spaces has been extensively used in philosophy, e.g.\ to study metaphors and vagueness \cite{douven2013vagueness}, and in psychology, e.g.\ to study perception in domains such as color \cite{DBLP:conf/amsterdam/Jager09} and music \cite{forth2010unifying}. However, the lack of automated methods for learning conceptual spaces from data has held back its adoption in the field of artificial intelligence. While a number of such methods have recently been proposed \cite{DBLP:conf/sigir/JameelBS17}, an important remaining problem is that these methods typically do not explicitly model concepts, i.e.\ they only learn the representations of the objects, while it is the concept representations that are mostly needed in applications such as knowledge base completion. The problem we study in this paper is to induce these missing concept representations from these object representations.

\begin{figure}
    \centering
            \includegraphics[width=0.95\linewidth]{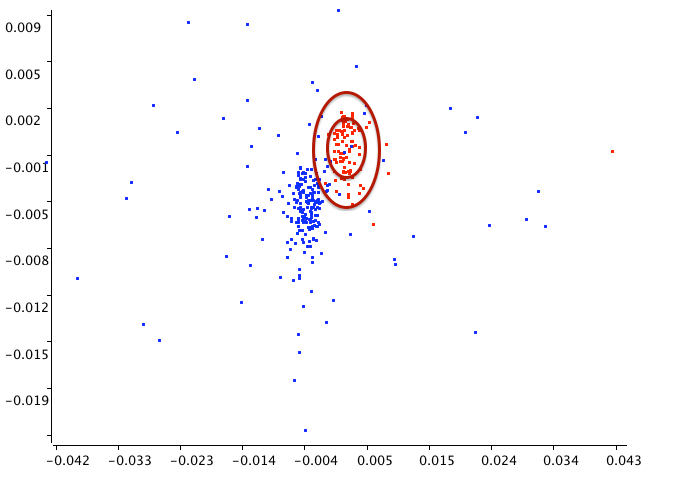}
    \caption{Two dimensions from a vector space embedding of places. Places known by SUMO or Wikidata to be train stations are shown in red.}
    \label{figConceptualSpacePlot}
\end{figure}

To illustrate the considered problem, Figure \ref{figConceptualSpacePlot} shows two dimensions from a higher-dimensional conceptual space of places.
The red dots correspond to places which are asserted to be train stations in the SUMO ontology or on WikiData. From this red point cloud, we can learn a soft boundary for the concept \emph{train station}, which is illustrated by the ellipsoidal contours in the figure. We can then plausible conclude that points which are within these boundaries are likely to be train stations.
In accordance with prototype theory, this model essentially assumes that the likelihood that an object is considered to be an instance of a train station depends on its distance to a prototype. Note that by considering ellipsoidal rather than spherical contours, we can take into consideration that different dimensions may have different levels of importance in any given context. In principle, there are several strategies that could be used to find suitable ellipsoids for a given concept, e.g.\ we could train a support vector machine with a quadratic kernel or we could fit a Gaussian distribution. The key problem with these methods, however, is that conceptual spaces usually have hundreds of dimensions, whereas only a few instances of each concept may be available, making learned concept representations potentially unreliable.

While this cannot be avoided in general, in many applications we have some background knowledge about how the considered concepts are interrelated. In this paper, we propose a Bayesian model which exploits such background knowledge while jointly learning a representations for all concepts. In particular, we assume that concepts can be modelled using Gaussians, and we use the available background knowledge to construct informative priors on the parameters of these Gaussians. We will consider two kinds of such background knowledge. First, we will consider logical dependencies between the concepts, which can be encoded using description logic (DL). For instance, SUMO encodes the knowledge that each instance of \textit{TrainStation} is an instance of \textit{terminalBuilding} or of \textit{transitBuilding}, which we should intuitively be able to exploit when learning representations for these concepts. Second, we will use the fact that many concepts themselves also correspond to objects in some conceptual space. For example, while train station is a concept in a conceptual space of \textit{places}, it is an object in a conceptual space of \textit{place types}. As we will see below, these representations of concepts-as-points can be used to implement a form of analogical reasoning. We experimentally demonstrate the effectiveness of our proposed model in a knowledge base completion task. Specifically, we consider the problem of identifying missing instances of the concepts from a given ontology. We show that our method is able to find such instances, even for concepts for which initially only very instances were known, or even none at all.

\section{Related Work}

\textbf{Learning Conceptual Spaces} 
One common strategy to obtain conceptual spaces is to learn them from human similarity judgments using multidimensional scaling. Clearly, however, such a strategy is only feasible in small domains. To enable larger-scale application, a number of approaches have recently been proposed for learning Euclidean\footnote{While the use of Euclidean spaces is quite natural, another common choice is to use a two-level representation, where a concept at the top level is a weighted set of properties, each of which corresponds to a convex region in a different Euclidean space. Such representations open up interesting possibilities, but there are currently no established methods for learning them in an automated way.} conceptual space representations in a purely data-driven way. In our experiments, we will in particular rely on the MEmbER model from \cite{DBLP:conf/sigir/JameelBS17}, which learns vector space representations that can be seen as approximate conceptual spaces. For instance, in contrast to most other vector space models, objects of the same semantic type are grouped in lower-dimensional subspaces within which dimensions corresponding to salient features (i.e.\ quality dimensions) can be found. In other words, this approach can be seen as learning a set of conceptual spaces, one for each considered semantic type, which are themselves embedded in a higher-dimensional vector space. Most importantly for this paper, the objective of the MEmbER model directly imposes the requirement that all entities which are strongly related to a given word (i.e.\ whose textual descriptions contain sufficiently many occurrences of the word) should be located within some ellipsoidal region of the space. It thus aims to learn a representation in which concepts can be faithfully modelled as densities with ellipsoidal contours, such as Gaussians.

Throughout this paper, we will assume that natural concepts can be modelled as (scaled) Gaussians. This corresponds to a common implementation of prototype theory \cite{rosch1973natural}, in which the prototype of a concept is represented as a point and the membership degree of an object is proportional to its exponential squared distance to the prototype. Note that, in general, prototypes do not have to be modelled as points, e.g.\ a more general approach is to model prototypes as regions \cite{douven2013vagueness}. However, the restriction to prototype points is a useful simplifying assumption if we want to learn reasonable concept representations from small numbers of instances. Similarly, while in principle it would be useful to model concepts as Gaussian mixture models, a strategy which was proposed in \cite{rosseel2002mixture} to generalize both prototype and exemplar models, this would only be feasible if a large number of instances of each concept were known.

\smallskip\noindent \textbf{Knowledge Graph Completion}
The main application task considered in this paper is knowledge base completion, i.e.\ identifying plausible facts which are missing from a given knowledge base. Broadly speaking, three types of highly complementary methods have been considered for this task. First, some methods focus on identifying and exploiting statistical regularities in the given knowledge base, e.g.\ by learning predictive latent clusters of predicates \cite{Kok:2007:SPI:1273496.1273551,DBLP:conf/akbc/RocktaschelR16,DBLP:conf/ilp/SourekMZSK16} or by embedding predicates and entities in a low-dimensional vector space \cite{NIPS20135071}. The second class consists of approaches which extract facts that are asserted in a text corpus. For example, starting with \cite{hearst1992automatic}, a large number of methods for learning taxonomies from text have been proposed \cite{kozareva2010semi,DBLP:conf/ijcai/AlfaroneD15}. Several authors have proposed methods that use a given incomplete knowledge base as a form of distant supervision, to learn how to extract specific types of fine-grained semantic relationships from a text corpus \cite{mintz2009distant,DBLP:conf/pkdd/RiedelYM10}. Thirdly, some methods, including ours, aim to explicitly represent concepts in some underlying feature space. For example, \cite{DBLP:conf/naacl/NeelakantanC15} represents each Freebase entity using a combination of features derived from Freebase itself and from Wikipedia, and then uses a max-margin model to identify missing types. In \cite{DBLP:conf/aaai/BouraouiJS17}, description logic concepts were modelled as Gaussians in a vector space embedding. Crucially, these existing works consider each concept in isolation, which requires that large numbers of instances are known, and this is often not the case.

\smallskip\noindent \textbf{Few Shot Learning}
Considerable attention has also been paid to the problem of learning categories for which no, or only few training examples are available, especially within the area of image recognition. For example, in one common setting, each category is defined w.r.t.\ a set of features, and the assumption is that we have training examples for some of the categories, but not for all of them. Broadly speaking, the aim is then to learn a model of the individual features, rather than the categories, which then makes it possible to make predictions about previously unseen categories \cite{palatucci2009zero,romera2015embarrassingly}. Other approaches instead exploit the representation of the category names in a word embedding \cite{socher2013zero}. We will similarly exploit vector space representations of concept names.

\section{Background}
We will rely on description logic encoding of how different concepts are related, and we will use a Bayesian approach for estimating Gaussians modelling these concepts. In this section, we briefly recall the required technical background on these two topics. For a more comprehensive discussion, we refer to \cite{Baader:2003:DLH:885746} and \cite{conjugateGaussian} respectively.

\smallskip\noindent \textbf{Description Logics}
Description logics are a family of logics which are aimed at formalizing ontological knowledge about the concepts from a given domain of interest. The basic ingredients are individuals, concepts and roles, which at the semantic level respectively correspond to objects, sets of objects, and binary relations between objects. A knowledge base in this context consists of two parts: a TBox, which encodes how the different concepts and roles from the ontology are related, and an ABox, which enumerates some of the instances of the considered concepts and roles. The TBox is encoded as a set of concept inclusion axioms of the form $C \sqsubseteq D$, which intuitively encodes that every instance of the concept $C$ is also an instance of the concept $D$. Here, $C$ and $D$ are so-called concept expressions. These concept expressions are either atomic concepts or complex concepts. In this paper we will consider complex concepts that are constructed in the following ways:
\begin{itemize}
\item If $C$ and $D$ are concept expressions, then $C \sqcap D$ and $C \sqcup D$ are also concept expressions, modelling the intersection and union of the concepts $C$ and $D$ respectively.
\item If $C$ is a concept expression and $R$ is a role, then $\exists R . C$ and $\forall R . C$ are also concept expressions. Intuitively, an individual belongs to the concept $\exists R . C$ if it is related (w.r.t.\ the role $R$) to some instance from $C$; an individual belongs to  $\forall R . C$ if it can only be related (w.r.t.\ $R$) to instances of $C$.
\end{itemize}

From a given knowledge base, we can typically infer further concept inclusion axioms and ABox assertions, although the complexity of such reasoning tasks crucially depends on the specific description logic variant that is considered (e.g.\ which types of constructs are allowed and what restrictions are imposed on concept inclusion axioms). Note that the methods we propose in this paper are independent of any particular description logic variant; we will simply assume that an external reasoner is available to infer such axioms.
 
\smallskip\noindent \textbf{Bayesian Estimation of Gaussians}
Suppose a set of data points $x_1,...,x_n \in \mathbb{R}$ have been generated from a univariate Gaussian distribution $G$ with a known variance $\sigma$ and an unknown mean. Rather than estimating a single value $\mu^*$ of this mean, in the Bayesian setting we estimate a probability distribution $M$ over possible means. Suppose our prior beliefs about the mean are modelled by the Gaussian $P = \mathcal{N}(\mu_P,\sigma^2_P)$. After observing the data points $x_1,...,x_n$ our beliefs about $\mu$ are then modelled by the distribution $M$ defined by:
\begin{align}\label{eqSamplingMu}
M(\mu) \propto p(x_1,...,x_n \,|\, \mu,\sigma^2) \cdot P(\mu)
\end{align}
It can be shown that this distribution $M$ is a Gaussian $\mathcal{N}(\mu_M,\sigma_M^2)$, where:
\begin{align*}
\sigma_M^2 &= \frac{\sigma^2 \sigma_P^2}{n\sigma_P^2 + \sigma^2} &
\mu_M &= \sigma_M^2 \left(\frac{\mu_P}{\sigma_P^2} + \frac{\sum_i x_i}{\sigma^2} \right)
\end{align*}
Now consider a setting where the variance of the Gaussian is unknown, but the mean is known to be $\mu$. For computational reasons, prior beliefs on the variance are usually modelled using an inverse $\chi^2$ distribution (or a related distribution such as inverse Gamma); let us write this as $Q=\chi^{-2}(\nu_Q,\sigma_Q^2)$. Intuitively, this means that we a priori believe the variance is approximately $\sigma_Q^2$, with $\nu_Q$ expressing the strength of this belief. After observing the data, our beliefs about the possible values of $\sigma^2$ are modelled by the distribution $S$, defined by:
\begin{align}\label{eqSamplingSigma}
S(\sigma^2) \propto p(x_1,...,x_n \,|\, \mu,\sigma^2) \cdot P(\sigma^2)
\end{align}

\noindent It can be shown that $S = \chi^{-2}(\nu_S,\sigma_S^2)$ where:
\begin{align*}
\nu_S &= \nu_Q + n &
\sigma^2_S &= \frac{\nu_Q \sigma_Q^2 + \sum_i (x_i-\mu)^2}{\nu_S}
\end{align*}

\section{Learning Concept Representations}
We assume that a description logic ontology is given, and that for each individual $a$ mentioned in the ABox of this ontology, a vector representation $v_a \in\mathbb{R}^n$ is available. For our experiments, these representations will be obtained using the MEmbER model, although in principle other vector space models could also be used. The task we consider is to learn a Gaussian $G_C=(\mu_C,\Sigma_C)$ for each concept which is mentioned in the TBox or ABox, as well as for all constituents of these concepts (e.g.\ if the concept $C_1 \sqcup... \sqcup C_k$ is mentioned then we also learn Gaussians for $C_1,...,C_k$), along with a scaling factor $\lambda_C>0$, such that the probability that the individual $a$ belongs to concept $C$ is given by:
\begin{align}\label{eqProbConcept}
P(C | v_a) &= \lambda_C \cdot G_{C}(v_a)
\end{align}
Intuitively, the variance of $G_C$ encodes how much the instances of $C$ are dispersed across the space, while $\lambda_C$ allows us to control how common such instances are. Formally, if we assume that the prior on $v_a$ is uniform, $\lambda_C$ is proportional to the prior probability $P(C)$ that an individual belongs to $C$.

Given that the number of known instances of the concept $C$ might be far lower than the number of dimensions $n$, it is impossible to reliably learn the covariance matrix $\Sigma_C$ without imposing some drastic regularity assumptions. To this end, we will make the common simplifying assumption that $\Sigma_C$ is a diagonal matrix. The problem of estimating the multivariate Gaussian $G_{C}$ then simplifies to the problem of estimating $n$ univariate Gaussians. In the following, for a multivariate Gaussian $G=\mathcal{N}(\mu_G,\Sigma_G)$, we write $\mu_{G,i}$ for the $i^{\textit{th}}$ component of $\mu_{G}$ and  $\sigma_{G,i}^2$ for the $i^{\textit{th}}$ diagonal element of $\Sigma_{G}$.

To find the parameters of these Gaussians, we will exploit background knowledge about the logical relationships between the concepts. However, this means that the parameters of the Gaussian corresponding to some concept $C$ may depend on the parameters of the Gaussians corresponding to other concepts. To cope with the fact that this may result in cyclic dependencies, we will rely on Gibbs sampling, which is explained next. This process will crucially rely on the construction of informative priors on the parameters of the Gaussians, which is discussed in Sections \ref{secPriorMean} and \ref{secPriorVariance}. Finally, Section \ref{secScalingFactor} will explain how the scaling factors $\lambda_C$ are estimated.

\subsection{Gibbs Sampling}
The purpose of Gibbs sampling is to generate sequences of parameters $\mu_C^0,\mu_C^1,...$ and $\Sigma_C^0,\Sigma_C^1,...$ for each concept. To make predictions, we will then average over the samples in these sequences. We will write $\mu_{C,i,j}$ for the $i^{\textit{th}}$ component of $\mu_C^j$ and $\sigma_{C,i,j}^2$ for the $i^{\textit{th}}$ diagonal element of $\Sigma_C^j$. 

The initial parameters $\mu_C^0$ and $\Sigma_C^0$ are chosen as follows. If $v_1,...,v_k$ are the vector representations of the known instances of $C$ and $k\geq 2$, we choose:
\begin{align*}
\mu_C^0 &= \frac{1}{k}\sum_l v_l &
\sigma_{C,i,0}^2 &= \frac{1}{k-1}\sum_l (v_{l,i}-\mu_{C,i,0})^2
\end{align*}
where we write $v_{l,i}$ for the $i^{\textit{th}}$ coordinate of $v_l$. If $k\leq 1$ the parameters $\mu_C^0$ and $\sigma_{C,i,0}^2$ are estimated based on the superconcepts of $C$ in the ontology; more details about these corner cases are provided in the appendix.

After the initial parameters have been chosen, we repeatedly iterate over all concepts. In the $i^{\textit{th}}$ iteration ($i>0$), we choose the next samples  $\mu_C^{i}$ and $\Sigma_C^{i}$ for each concept $C$, according to \eqref{eqSamplingMu} and \eqref{eqSamplingSigma} respectively. To do this, however, we first need to define prior probabilities on $\Sigma_C^{i}$ and $\mu_C^{i}$. These prior probabilities will be constructed by taking into account the available background knowledge about how the different concepts are interrelated, as we explain in detail in Sections \ref{secPriorMean} and \ref{secPriorVariance}. In particular, the prior probabilities on  $\Sigma_C^{i}$ and $\mu_C^{i}$ will be defined in function of the parameters of the Gaussians of the other concepts. When using Gibbs sampling, we always use the most recent samples of the parameters of these other concepts. 

For the ease of presentation, we will write $\mu_B^{*}$ and $\Sigma_B^{*}$ for the most recent samples of $\mu_B$ and $\Sigma_B$. In other words, $\mu_B^{*}=\mu_B^{i-1}$ or $\mu_B^{*}=\mu_B^{i}$ holds, depending on whether $\mu_B$ was already updated in the current iteration of the Gibbs sampler, and similar for $\Sigma_B^{*}$. We also write $G^*_B$ for $\mathcal{N}(\mu_B^*,\Sigma_B^*)$, i.e.\ the most recent estimation of the Gaussian $G_B$. 
Similarly, we also use the notations $\mu_{B,i,*}$ and $\sigma_{B,i,*}^2$ to refer to the $i^{\textit{th}}$ component of $\mu_B^{*}$ and the $i^{\textit{th}}$ diagonal element of $\Sigma_B^{*}$ respectively.

\subsection{Priors on the Mean}\label{secPriorMean}
The type of information that is available to construct a prior on the mean $\mu_C^i$ is different for atomic and for complex concepts, which is why we discuss these cases separately. 

\subsubsection{Atomic Concepts}
For an atomic concept $A$, we use two types of information to construct the prior $P_A = \mathcal{N}(\mu_{P_A},\Sigma_{P_A})$ that is used for sampling $\mu_A^i$. First, the TBox may contain a number of axioms of the form $A\sqsubseteq C_1,...,A\sqsubseteq C_k$.
If $A\sqsubseteq C_l$ holds then $\mu_A$ should correspond to a plausible instance of $C_l$. In particular, we would expect the probability $G^*_{C_l}(\mu_A)$ to be high.

Second, if a vector representation $v_A$ of the concept $A$ itself is available, it can also provide us with useful information about the likely values of $\mu_A$. Suppose $B_1,...,B_r$ are atomic concepts such that the TBox contains or implies the axioms $B_1 \sqsubseteq C_l,...,B_r \sqsubseteq C_l$ (in addition to $A\sqsubseteq C_l$). We will refer to $B_1,...,B_r$ as the siblings of $A$ w.r.t.\ $C_l$. The information that we want to encode in the prior $P_A$ is that the vector differences $v_{B_1} - \mu_{B_1}^*,...,v_{B_r} - \mu_{B_r}^*$ should all be similar to the vector difference $v_A - \mu_{A}$. This is motivated by the fact that, in the context of word embeddings, analogical word pairs typically have similar vector differences \cite{mikolov2013linguistic,Vylomova2016}. In particular, it corresponds to the intuitive assumption that the relation between the prototype of a concept and the vector space embedding of the concept name should be analogous for all concepts, and in particular for all siblings of $A$. This intuition can be encoded by estimating a Gaussian $E_{C_l}=\mathcal{N}(\mu_{E_{C_l}},\Sigma_{E_{C_l}})$ from the vector differences $v_{B_1} - \mu_{B_1}^*,...,v_{B_r} - \mu_{B_r}^*$ and the representations $v_{B_1},...,v_{B_r}$ themselves, as follows (assuming $r\geq 2$):
\begin{align*}
\mu_{E_{C_l}} &= v_A + \frac{1}{r}\sum_{u=1}^r (\mu_{B_u}^* - v_{B_u})\\
\sigma_{E_{C_l},j}^2 &= \frac{1}{r-1} \sum_{u=1}^r (v_{A,j} + \mu_{B_u,j}^* - v_{B_u,j} - \mu_{E_{C_l},j}) ^2
\end{align*}
where $v_{A,j}$ is the $j^{th}$ coordinate of $v_A$, and similar for $v_{B_u,j}$. Using this Gaussian, we can encode our intuition by requiring that $E_{C_l}(\mu_A)$ should be high. If $r=1$, then $\sigma_{E_{C_l},j}^2$ is estimated based on the superconcepts of $C_l$; details about this corner case can be found in the online appendix.

Combining both types of background knowledge, we choose a prior $P_A$ which encodes that $P_A(\mu_A)$ is proportional to $G_{C_1}^*(\mu_A)\cdot ...G_{C_l}^*(\mu_A) \cdot E_{C_1}(\mu_A)\cdot ... \cdot E_{C_l}(\mu_A)$, as follows:
\begin{align*}
\sigma_{P_A,j}^2 &= \left(\sum_{u=1}^k \frac{1}{\sigma_{C_u,j,*}^{2}} + \frac{1}{\sigma_{E_{C_u},j}^{2}}\right)^{-1}\\
\mu_{P_A,j} &= \sigma_{P_A,j}^2 \sum_{u=1}^k \left( \frac{\mu_{C_u,j,*}}{\sigma_{C_u,j,*}^{2}} + \frac{\mu_{E_{C_u},j}}{\sigma_{E_{C_u},j}^{2}}\right)
\end{align*}
For the ease of presentation, here we have assumed that $A$ has at least one sibling w.r.t.\ each $C_l$. In practice, if this is not the case, the corresponding Gaussian $E_{C_l}(\mu_A)$ is simply omitted. Note that when the translation assumption underlying the Gaussians $E_{C_l}$ is not satisfied, the associated variances $\sigma_{E_{C_u},j}^{2}$ will be large, and accordingly the information from the vector space embeddings will be mostly ignored.

\subsubsection{Complex Concepts}
To construct a prior on $\mu_C^i$ for a complex concept $C$, we can again use concept inclusion axioms of the form $C\sqsubseteq C_1,...,C\sqsubseteq C_k$ entailed from the ontology, but for complex concepts we do not have access to a vector space representations. However, additional prior information can be derived from the atomic concepts and roles from which $C$ is constructed. For example, let $C \equiv D_1 \sqcap ... \sqcap D_s$. Then we can make the assumption that $P(C|v)$ is proportional to $G_{D_1}^*(c)\cdot ...\cdot G_{D_s}^*(c)$. The product of these Gaussians is proportional to a Gaussian $H_C^*$ with the following parameters:
\begin{align*}
\sigma_{H_C,j}^2 &= \left(\sum_{i=1}^s \frac{1}{\sigma_{G_{D_i},j,*}^{2}}\right)^{-1}\\
\mu_{H_C,j} &= \sigma_{H_C,j}^2 \sum_{i=1}^s \frac{\mu_{G_{D_i},j,*}}{\sigma_{G_{D_i},j,*}^{2}}
\end{align*}
This leads to the following choice for the prior:
\begin{align*}
\sigma_{P_C,j}^2 &= \left(\frac{1}{\sigma_{H_C,j,*}^{2}} + \sum_{u=1}^k \frac{1}{\sigma_{C_u,j,*}^{2}}\right)^{-1}\\
\mu_{P_C,j} &= \sigma_{P_C,j}^2\left(\frac{\mu_{H_C,j,*}}{\sigma_{H_C,j,*}^{2}} +  \sum_{u=1}^k \frac{\mu_{C_u,j,*}}{\sigma_{C_u,j,*}^{2}} \right)
\end{align*}\\
For complex concepts of the form $D_1\sqcup ... \sqcup D_s$, $\exists R\,.\,C$ and $\forall R\,.\,C$, a similar strategy can be used. Details about these cases can be found in the online appendix.
\subsection{Priors on the Variance}\label{secPriorVariance}

For a concept $C$, we write $Q_{C,j}=\chi^{-2}(\nu_{Q_{C,j}},\sigma_{Q_{C,j}}^2)$ for the prior on $\sigma^2_{C,j}$, the $j^{\textit{th}}$ diagonal element of $\Sigma_C$. We now discuss how the parameters $\nu_{Q_{C,j}}$ and $\sigma_{Q_{C,j}}^2$ are chosen.

\subsubsection{Atomic Concepts}
Let $A$ be an atomic concept. We will exploit two types of information about the variance $\sigma^2_{A,j}$. First, if $A\sqsubseteq C$ then $\sigma_{A,j}^2 \leq \sigma_{C,j}^2$ should hold. Second, we might expect that the covariance matrix of $G_A$ is similar to that of the most closely related concepts. Specifically, let $B_1,...,B_k$ be all the atomic siblings of $A$ (i.e.\ each $B_l$ is an atomic concept, and there is some $C$ such that both $A\sqsubseteq C$ and $B_l\sqsubseteq C$ appear in the TBox). Then one possibility is to choose $\sigma_{Q_{C,j}}^2$ as the average of $\sigma_{B_1,j}^2,...,\sigma_{B_k,j}^2$. If we have a vector space representation for $A$ and for some of its (atomic) siblings, then we can improve on this estimation by only considering the most similar siblings, i.e.\ the siblings whose vector space representation is closest in terms of Euclidean distance. In particular, let $\mathcal{B}\subseteq \{B_1,...,B_k\}$ be the set of the $\kappa$ most similar siblings of $A$, and let $C_1,...,C_l$ be the set of concepts for which the TBox contains the axiom $A\sqsubseteq C_l$, then we choose:
$$
\sigma_{Q_{A,j}}^2 = \min\left(\min_l \sigma_{C_l,j,*}^2, \frac{1}{\kappa}\sum_{B\in \mathcal{B}} \sigma_{B,j,*}^2\right)
$$
The parameter $\nu_{Q_{C,j}}$ intuitively reflects how strongly we want to impose the prior on $\sigma^2_{A,j}$. Given that even closely related concepts could have a considerably different variance, we set $\nu_{Q_{A,j}}$ as a small constant $\eta$ for each $A$ and $j$.

\subsubsection{Complex Concepts}
For a complex concept $C$, the covariance matrix of $H_{C}^*$, with $H_{C}^*$ the Gaussian constructed for complex concepts in Section \ref{secPriorMean}, can be used to define the prior on $\sigma^2_{C,j}$. Let $C_1,...,C_k$ be the concepts for which the TBox contains or implies the axiom $C\sqsubseteq C_l$ then we choose:

$$
\sigma_{Q_{C,j}}^2 = \min(\min_l \sigma_{C_l,j,*}^2, \sigma_{H_C,j,*}^2)
$$
Furthermore, we again set $\nu_{Q_{C,j}}=\eta$.

\subsection{Making Predictions}\label{secScalingFactor}

For a given individual with vector representation $v$, we can estimate the probability $P(C|v)$ that this individual is an instance of concept $C$ as follows:
\begin{align}\label{eqDefConceptProbAveraged}
P(C|v) =  \frac{\lambda_C }{N} \sum_{i=1}^N p(v; \mu_{C}^i,\Sigma_C^i)
\end{align}
Note that compared to \eqref{eqProbConcept}, in \eqref{eqDefConceptProbAveraged} the parameters of $G_C$ are averaged over the Gibbs samples. As usual with Gibbs sampling, the first few samples are discarded, so here $\mu_{C}^1$ and $\Sigma_C^1$ refer to the first samples after the burn-in period. 

The number of Gibbs sample that we used is equal to 1000 where each sample is generated after 25 every 25 iterations. The burn-in period that we use is fixed to 200 samples.

We estimate the scaling factor $\lambda_C$ by maximizing the likelihood of the training data. In particular, let $v_1,...,v_s$ be the vector representations of the known instances of $C$, and let $u_1,...,u_r$ be the vector representations of the individuals which are not asserted to belong to $C$. Then we choose the value of $\lambda_C$ that maximizes the following expression:
$$
\sum_{i=1}^s \log(\lambda_C P(v_i |C)) + \sum_{i=1}^r \log(1-\lambda_C P(u_i |C))
$$
which is equivalent to maximizing:
\begin{align}\label{eqLossScaling}
s \lambda_C + \sum_{i=1}^r \log(1-\lambda_C P(u_i |C))
\end{align}
Note that for concepts without any known instances, we would obtain $\lambda_C=0$, which is too drastic. To avoid this issue, we replace $s$ by $s+1$ in \eqref{eqLossScaling}, which is similar in spirit to the use of Laplace smoothing when estimating probabilities from sparse frequency counts. Furthermore note that the estimation of $\lambda_C$ relies on a closed world assumption, i.e.\ we implicitly assume that individuals do not belong to $C$ if they are not asserted to belong to $C$. Since this assumption may not be correct (i.e.\ some of the individuals $u_1,...,u_r$ may actually be instances of $C$, even if they are not asserted to be so), the value of $\lambda_C$ we may end up with could be too low. This is typically not a problem, however, since it simply means that the predictions we make might be more cautious then they need to be. 


\section{Experimental Results}

\begin{table*}[t]
\centering
\footnotesize
\begin{tabular}{|@{\hspace{0.1pt}}c@{\hspace{.1pt}}|cccc|cccc|cccc|}
\hline 
 & \multicolumn{4}{c|}{SVM-Linear} &  \multicolumn{4}{c|}{SVM-Quad} & \multicolumn{4}{c|}{Gibbs} \tabularnewline
\hline 
 & Pr & Rec & F1 & AP & Pr & Rec & F1 & AP & Pr & Rec & F1 & AP  \tabularnewline
\hline 
$1 \leq |X|\leq 5$ &0.033 &0.509 &0.062 &0.055    &0.086 &0.046 &0.060& 0.144  &0.258 &0.508 &0.343 &0.328   \tabularnewline
$5 < |X|\leq 10$   &0.084&0.922 &0.154 &0.067    &0.116&0.404&0.180& 0.163     &0.202& 0.474& 0.283& 0.340     \tabularnewline
$10< |X|\leq 50$   &0.111&0.948& 0.199 & 0.081   &0.151& 0.382& 0.216 & 0.247   &0.242 & 0.886 & 0.380 & 0.276    \tabularnewline
$|X|>50$           &0.153& 0.217&0.180&0.230   &0.224& 0.721 & 0.342 & 0.260  &0.361& 0.678 & 0.471 & 0.404  \tabularnewline
\hline 
\end{tabular}
\caption{Results of the proposed model and the baselines.}
\label{set1}
\end{table*}

\begin{table*}[t]
\centering
\footnotesize
\begin{tabular}{|@{\hspace{0.1pt}}c@{\hspace{.1pt}}|cccc|cccc|cccc|}
\hline 
 & \multicolumn{4}{c|}{Gibbs-flat} & \multicolumn{4}{c|}{Gibbs-emb} & \multicolumn{4}{c|}{Gibbs-DL}\tabularnewline
\hline 
 &Pr & Rec & F1 & AP &  Pr & Rec & F1 & AP &  Pr & Rec & F1 & AP\tabularnewline
\hline 
$1 \leq |X|\leq 5$  &  0.212 &  0.416    &0.281 & 0.290   &  0.201 & 0.540  &0.293 & 0.262  &  0.226 &  0,498  &0.311 & 0.304   \tabularnewline
$5 < |X|\leq 10$    &  0.186  &  0.368   &0.247 & 0.273   & 0.173 & 0.357   &0.233 & 0.262  &  0.417 & 0.192 &0.263 &0.328  \tabularnewline
$10< |X|\leq 50$    & 0.199  &   0.496   &0.284 & 0.210   & 0.207  &  0.513 &0.295 & 0.233  &  0.218 & 0.670 &0.329 & 0.251  \tabularnewline
$|X|>50$            & 0.316  &  0.312    &0.314 & 0.328   &  0.321 & 0.373  &0.345 &0.321   &  0.344 &0.450 &0.390 & 0.369  \tabularnewline
\hline 
\end{tabular}
\caption{Results for the variants of the proposed model.}
\label{set1}
\end{table*}

In this section, we experimentally evaluate our method against a number of baseline methods. In our experiments, we have used SUMO\footnote{\url{http://www.adampease.org/OP/}}, which is a large open-domain ontology. An important advantage of using SUMO is that several of its concepts and individuals are explicitly mapped to WordNet, which itself is linked to WikiData. This means that we can straightforwardly align this ontology with our entity embedding. For concepts and individuals for which we do not have such a mapping, we use BabelNet\footnote{We have used the BabelNet Java API, which is available at \url{http://babelnet.org}} to suggest likely matches. The SUMO ontology contains 4558 concepts, 778 roles and 86475 individuals.

We split the set of individuals into a training set $I_{\textit{train}}$ containing 2/3 of all individuals, and a test set $I_{\textit{test}}$ containing the remaining 1/3. All ABox assertions involving individuals from $I_{\textit{train}}$ are used as training data. 
The considered evaluation task is to decide for a given assertion $A(a)$ (meaning ``$a$ is an instance of $A$'') whether it is correct or not. As positive examples, we use all assertions from the ABox involving individuals from $I_{\textit{test}}$. To generate negative test examples, we use the following strategies. First, for each positive example $A(a)$ and each concept $B\neq A$ such that the TBox implies $A\sqsubseteq B$, we add a negative example by randomly selecting an individual $x$ such that $B(x)$ can be deduced from SUMO while $A(x)$ cannot. Second, for each positive example $A(a)$, we also add 10 negative examples by randomly selecting individuals among all those that are not known to be instances of $x$. Note that even if $A(x)$ is not asserted by SUMO, it may be the case that $x$ is an instance of $A$. This means that in a very small number of cases, the selected negative examples might actually be positive examples. The reported results are thus a lower bound on the actual performance of the different methods. Importantly, the relative performance of the different methods should not be affected by these false negatives.

The performance is reported in terms of average precision (AP), and micro-averaged precision (Pr), recall (Rec) and F1 score. To compute the AP scores, we rank the assertions from the test data (i.e.\ the correct ABox assertions as well as the constructed negative examples), across all considered concepts, according to how strongly we believe them to be correct, and then we compute the average precision of that ranking. To give a clearer picture of the performance of the different methods, however, we will break up the results according to the number of training examples we have for each concept (see below). 
Note that AP only evaluates our ability to rank individuals, and hence does not depend on the scaling factors $\lambda_A$. The precision, recall and F1 scores, however, do require us to make a hard choice. 

As baselines, we have considered a linear and quadratic support vector machine (SVM).
We will refer to our model as \emph{Gibbs}. We also consider three variants of our method: \emph{Gibbs-flat}, in which flat priors are used (i.e.\ no dependencies between concepts are taken into account), \emph{Gibbs-emb}, in which the priors on the mean and variance are only obtained from the embedding (i.e.\ no axioms from the TBox are taken into account), and \emph{Gibbs-DL}, in which the priors on the mean and variance are only obtained from the TBox axioms (i.e.\ the embedding is not taken into account).

The results are summarized in Table \ref{set1}, where $|X|$ refers to the number of training examples for concept $X$. Overall, our model consistently outperforms the baselines in both F1 and MAP score. For concepts with few known instances, the gains are substantial. Somewhat surprisingly, however, we even see clear gains for larger concepts.  Regarding the variants of our model, it can be observed that using TBox axioms to estimate the priors on the mean and variance (\emph{Gibbs-DL}) leads to better results than when a flat prior is used. The model \emph{Gibbs-emb}, however, does not outperform \emph{Gibbs-flat}. This means that the usefulness of the embedding, on its own, is limited. However, the full model, where the embedding is combined with the TBox axioms, does perform better than \emph{Gibbs-DL}.

\section{Conclusions and Future Work}

We have proposed a method for learning conceptual space representations of concepts. In particular,  we associate with each concept a Gaussian distribution over a learned vector space embedding, which is in accordance with some implementations of prototype theory. In contrast to previous work, we explicitly take into account known dependencies between the different concepts when estimating these Gaussians. To this end, we take advantage of description logic axioms and information derived from the vector space representation of the concept names. This means that we can often make faithful predictions even for concepts for which only a few known instances are specified.

\section*{Acknowledgments}
This work was supported by ERC Starting Grant 637277.

\ifappendix
\appendix

\section{Additional Technical Details}

\subsection{Initialization of the Gibbs Sampler}
When $C$ has less than two instances, the parameters $\sigma_{C,i,0}^2$ cannot be estimated from the instances of $C$ alone; if $C$ has no instances, then this also holds for the parameter $\mu_C^0$. In these cases, we initialize these parameters based on the parent concepts of $C$, which we define as follows. We call $B$ a parent of $C$ if the axiom $C\sqsubseteq B$ is included in the TBox or is included in the deductive closure. We will write $\textit{parents}(C)$ for the set of all parents of $C$, if $C$ has at least one parent; otherwise we define $\textit{parents}(C)=\{\top\}$. 

Let $k$ again be the number of known instances of the concept $C$. If $k=1$ we choose $\mu_C^0=v_1$, but we cannot estimate $\sigma_{C,i,0}^2$ from the instances of $C$. Therefore, we instead initialize $\sigma_{C,i,0}^2$ as the average of $\sigma_{B,i,0}^2$ over all $B$ in $\textit{parents}(C)$. Similarly, if $k=0$ then both $\mu_C^0$ and $\sigma_{C,i,0}^2$ are chosen by averaging these parameters over the parents of $C$. Note that because we only rely on the parents of $C$, there can be no cyclic dependencies.

\subsection{Priors on the Mean of Atomic Concepts}

\subsubsection{Atomic Concepts}
If $r=1$, then $\sigma_{E_{C_l},j}^2$ is estimated as the average of $\sigma_{E_{D},j}^2$ over each $D$ such that the TBox contains the axiom $C_l \sqsubseteq D$.

\subsection{Priors on the Mean of Complex Concepts}

\subsubsection{Modelling Unions}
Suppose $C = D_1 \sqcup ... \sqcup D_s$. One possibility would then be to model $C$ as a Gaussian mixture. However, in many cases, $D_1,...,D_s$ might intuitively correspond to neighboring concepts, in which case instances that are in between $D_1,...,D_S$ might be much more representative for $C$ than for any of the concepts $D_1,...,D_s$. For example, in the context of color grading, we might define $\textit{CoolColor} = \textit{Blue} \sqcup \textit{Green}$. A teal color might then be highly representative for $\textit{CoolColor}$, whereas it is a borderline case of both $\textit{Blue}$ and $\textit{Green}$. Accordingly, we model $H_C^*$ as a single Gaussian, whose parameters correspond to those of a Gaussian mixture of $G_{D_1}^*,...,G_{D_s}^*$ with uniform mixture weights:
\begin{align*}
\mu_{H_C}^* &= \frac{1}{s} \sum_i \mu_{G_{D_i}^*}\\
\sigma_{H_C,j,*}^2 &= \frac{1}{s} \left (\sum_i \sigma_{G_{D_i},j,*}^2 + (\mu_{G_{D_i},j,*})^2 \right) - \mu_{H_C,j,*}^2
\end{align*}

\subsubsection{Modelling Role Restrictions}
Modelling relations has not received as much attention as modelling concepts. However, within the context of word embeddings, it was found that many types of relations correspond to vector translations \cite{mikolov2013linguistic,Vylomova2016}. Accordingly, we will use the vector $v_b-v_a$ as a representation of the relationship between the individuals $a$ and $b$, and model roles from the ontology as Gaussians over such vector differences.  In particular, let $(h_1,t_1),...,(h_k,t_k)$ be the vector representations of the instances of $R$. Then we represent $R$ as the Gaussian  $G_R = (\mu_R,\Sigma_R)$ estimated (for $k\geq 2$) as follows:
\begin{align*}
\mu_R &= \frac{1}{k}\sum_l t_l-h_l\\
\sigma_{R,i}^2 &= \frac{1}{k-1}\sum_l (t_{l,i}-h_{l,i}-\mu_{R,i})^2
\end{align*}
If $k<2$ we cannot estimate the parameters of $G_R$ in a meaningful way, and we will not attempt to make any predictions about $R$. Accordingly, we will not consider role restrictions involving $R$ either. Furthermore, note that we will not use Gibbs sampling to estimate the parameters of the Gaussians modelling roles, as neither the ontology nor the entity embedding typically has any information about role dependencies that could easily be exploited.

Now assume that $C$ is of the form $\exists R. D$. If $v_a$ is the vector representation of an instance of $A$, then $A\sqsubseteq \exists R. D$ means that $v_A = v_B-v_R$ with $v_B$ and $v_R$ the vector representation of some instance of $B$ and some instance of $R$ respectively. A natural choice for $H_{\exists R. D}$ is thus to use the distribution of $X_B-X_R$, where $X_B$ and $X_R$ are random variables denoting an instance of $B$ and an instance of $R$ respectively. The random variable $X_B-X_R$ has a normal distribution with the following parameters:
\begin{align}
\mu_{H_C}^* &= \mu_{G_D^*} - \mu_{G_R} \label{eqExistsPriorMu1}\\
\sigma_{H_C,r,*}^2 &= \sigma_{G_D,r,*}^2 + \sigma_{G_R,r}^2 \label{eqExistsPriorMu2}
\end{align}
Finally, assume that $C$ is of the form $\forall R. B$. As the quantifiers $\exists$ and $\forall$ do not have a direct counterpart, we define $H_{\forall R.D}^*$ in the same way as $H_{\exists R.D}^*$.

\fi

\bibliography{commonsense,wordembedding,entity}
\bibliographystyle{named}

\end{document}